\documentclass[letterpaper, 10 pt, conference]{ieeeconf}  % Comment this line out if you need a4paper

\IEEEoverridecommandlockouts                              % This command is only needed if 
\makeatletter
\let\NAT@parse\undefined
\makeatother
\usepackage{cite}

\usepackage[linesnumbered, ruled, vlined]{algorithm2e}
\usepackage{graphicx} % for pdf, bitmapped graphics files
\usepackage{subcaption}
\usepackage[font=small]{caption}
\usepackage{stfloats}
\usepackage{booktabs}
\usepackage{times}
\usepackage{acronym}
\usepackage{soul}
\usepackage{xcolor}
\usepackage{censor}
\usepackage{verbatim}
\usepackage{makecell}
\usepackage{multirow}
\usepackage{amsmath} % assumes amsmath package installed
\usepackage{amssymb}  % assumes amsmath package installed

\usepackage{amsthm}
\newtheorem{theorem}{Theorem}
\usepackage{mathtools}
\usepackage{microtype}
\usepackage{hyperref}
\hypersetup{
    colorlinks=true,
    hidelinks=true
}
\usepackage{xspace}

\newcommand{\Glimpse}[1]{q(#1)}

\newcommand{\rtc}[1]{g(#1)}

\newcommand{\rtg}[1]{\hat{h}(#1)}

\newcommand{\asar}{A$^{\mathrm{SAR}}$\xspace}{}

\newcommand{\bel}{b}
\newcommand{\ibel}{\bar{b}}

\newenvironment{proof*}{\par\noindent\textit{Proof sketch.} }{\hfill$\square$\par}
\title{\LARGE \bf
A\textsuperscript{SAR}: $\varepsilon$-Optimal Graph Search for Minimum Expected-Detection-Time Paths with Path Budget Constraints for Search and Rescue (SAR)}

\author{Eric Mugford\textsuperscript{1} and Jonathan D.\ Gammell\textsuperscript{1}% <-this % stops a space
\thanks{\textsuperscript{1}Estimation, Search, and Planning (ESP) Research Group, Queen’s University, Kingston ON, Canada. {\tt\small \{19egm4,gammell\}@queensu.ca}}%
\thanks{The authors would like thank Ryan Berry, Grant Keefe, Jeremy Hayes, and Allan Taylor for their support and advice on this work as well as Prof.\ Sara Pérez-Carabaza for sharing her work. This work was supported by the Natural Sciences and Engineering Research Council of Canada (NSERC) [RGPIN-2024-06637] and Public Safety Canada's Search and Rescue New Initiatives Fund (SAR NIF).}%
}

\DeclareMathOperator*{\argmin}{arg\,min}
\newcommand{\squeezeWords}{\looseness=-1}
\newcommand{\squeezeLine}{\enlargethispage*{\baselineskip}\pagebreak}
\begin{document}
%\raggedbottom

\maketitle
\thispagestyle{empty}
\pagestyle{empty}

%%%%%%%%%%%%%%%%%%%%%%%%%%%%%%%%%%%%%%%%%%%%%%%%%%%%%%%%%%%%%%%%%%%%%%%%%%%%%%%%
\begin{abstract}
Searches are conducted to find missing persons and/or objects given uncertain information, imperfect observers and large search areas in Search and Rescue (SAR). In many scenarios, such as Maritime SAR, expected survival times are short and optimal search could increase the likelihood of success. This optimization problem is complex for nontrivial problems given its probabilistic nature. 

Stochastic optimization methods search large problems by nondeterministically sampling the space to reduce the effective size of the problem. This has been used in SAR planning to search otherwise intractably large problems but the stochastic nature provides no formal guarantees on the quality of solutions found in finite time. 

This paper instead presents \asar, an $\varepsilon$-optimal search algorithm for SAR planning. It calculates a heuristic to bound the search space and uses graph-search methods to find solutions that are formally guaranteed to be within a user-specified factor, $\varepsilon$, of the optimal solution. It finds better solutions faster than existing optimization approaches in operational simulations. It is also demonstrated with a real-world field trial on Lake Ontario, Canada, where it was used to locate a drifting manikin in only 150s. \squeezeWords

\end{abstract}

%%%%%%%%%%%%%%%%%%%%%%%%%%%%%%%%%%%%%%%%%%%%%%%%%%%%%%%%%%%%%%%%%%%%%%%%%%%%%%%%
\section{Introduction}
Search and rescue (SAR) operations are inherently resource intensive and may include land-, air-, and/or marine-based searches. Aerial searches are commonly conducted from manned aircraft, which are expensive to procure and operate \cite{hc_scpac_2017_report7}, limiting the number available. This limited availability means that aerial assets are often late to arrive on scene, especially in remote locations.

Unmanned aerial vehicles (UAVs) are an alternative to manned aircraft for SAR operations. UAVs can be launched quickly and can search an area faster than land- or marine-based assets without the traditional expenses of manned aircraft.  
Crewed aircraft also generally employ simple search patterns, such as parallel track, that can be reliably flown by pilots operating large, relatively unmanoeuvrable platforms \cite{GovCan_CAMSAR_2014}. UAVs are more manoeuvrable and can instead perform complex search patterns optimized for individual scenarios. 

A meaningful optimization objective for selecting search patterns (i.e., searcher paths) is expected detection time (i.e., mean time to detection; MTTD). The MTTD of a searcher path is the average time to first detect the target along the path for a given probability distribution of target locations. Minimizing MTTD may increase the probability of survival of a missing person since survival probability decreases with time, especially in cold or maritime environments.

Stochastic optimization algorithms can reduce MTTD given a starting position, searcher-path budget, prior target probability distribution and sensor model \cite{Lanillos_thesis_2013, Pérez-Carabaza_2018, Lanillos_Yañez-Zuluaga_Ruz_Besada-Portas_2013}. These algorithms use nondeterministic sampling to reduce the complexity of the path-constrained MTTD problem \cite{Trummel_Weisinger_1986}. This allows them to search otherwise intractably large searcher-path problems but provides no formal guarantees on the quality of solutions found in finite time. 

\begin{figure}[tbp]

    \begin{center}
        \includegraphics[width=\linewidth]{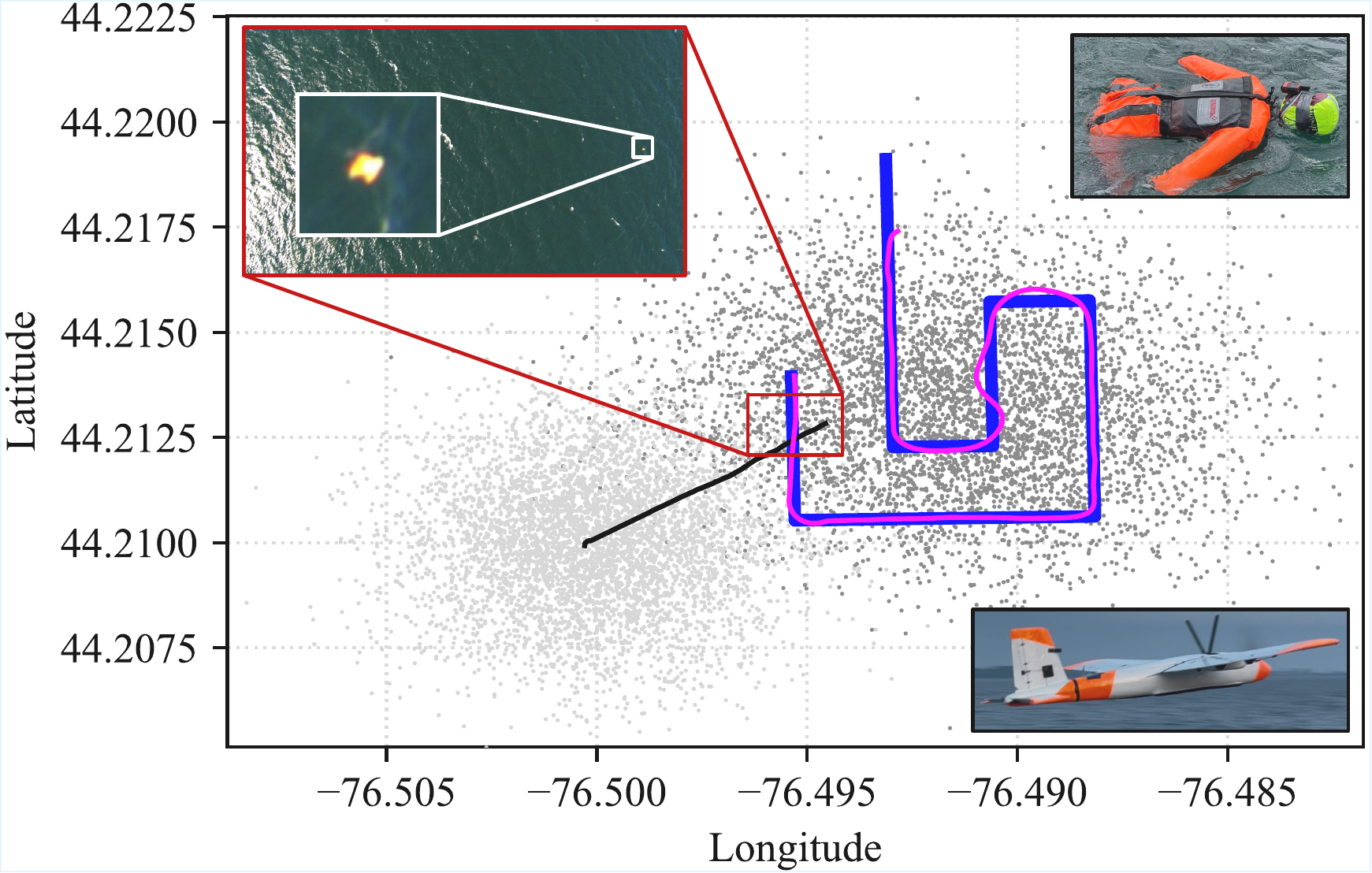}
    \end{center}
    \caption{\footnotesize Operational field experiment on Lake Ontario, Canada. A manikin (top right) drifted freely for one hour and was modelled with OpenDrift \cite{open_drift}, with initial particle positions in light grey and the predicted position at search time in dark grey. \asar computed the minimum MTTD path, which was flown by a fixed-wing UAV (bottom right). Planned (blue) and flown (magenta) paths are shown up to their intersection with the manikin at 150s. The UAV image of the manikin is shown top left and its true trajectory (black) was calculated after the experiment from an attached GPS tracker. \squeezeWords}
    \label{fig:place_holder}
\end{figure}

This paper instead presents \asar, an $\varepsilon$-optimal graph-search algorithm that can be applied to path-budget-constrained SAR planning. \asar bounds the search space by using an admissible heuristic which provides a lower bound on the MTTD objective function. It then uses a best-first search ordered by potential solution quality to find a solution that is within a user-defined suboptimality factor, $\varepsilon$, of the optimal path. This suboptimality factor allows the user to balance search speed and optimality guarantees.

\asar is tested in realistic maritime search scenarios generated using OpenDrift \cite{open_drift}, an open-source software package used by SAR professionals to model object drift. We show that \asar finds better solutions faster than a leading MTTD path planning method \cite{Pérez-Carabaza_2018} and traditional search patterns.  
The performance of \asar is also validated in a real-world experiment on Lake Ontario, Canada (Figure \ref{fig:place_holder}). \squeezeWords

\section{Related Work}
\begin{figure*}[t]      % “*” makes it double‐column; “t” places it at the top
  \centering
  %% Top row
  \begin{subfigure}[b]{0.225\linewidth}
    \includegraphics[width=\linewidth]{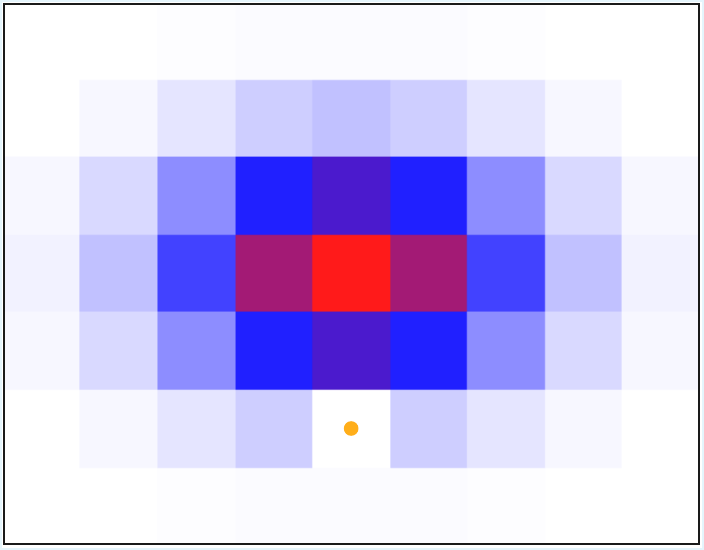}
    \label{fig:1}
    \caption{t = 1}
  \end{subfigure}
  \begin{subfigure}[b]{0.225\linewidth}
    \includegraphics[width=\linewidth]{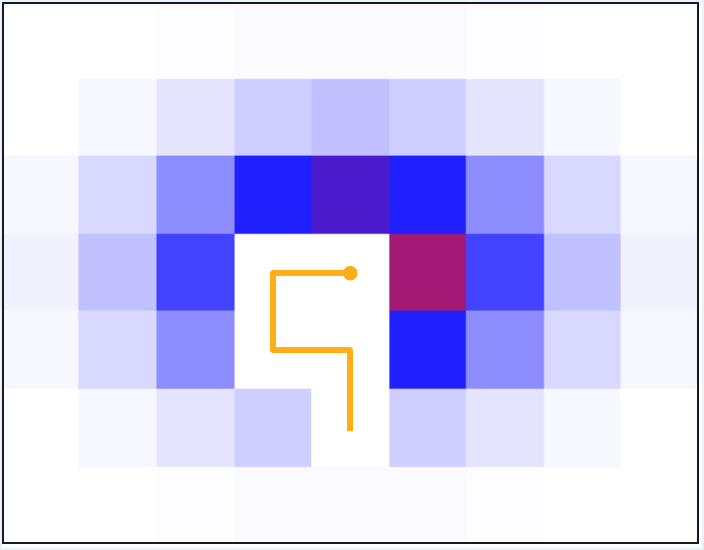}
    \label{fig:2}
    \caption{t = 5}
  \end{subfigure}
  %\vspace{1ex}         % small vertical gap between rows
  %% Bottom row
  \begin{subfigure}[b]{0.225\linewidth}
    \includegraphics[width=\linewidth]{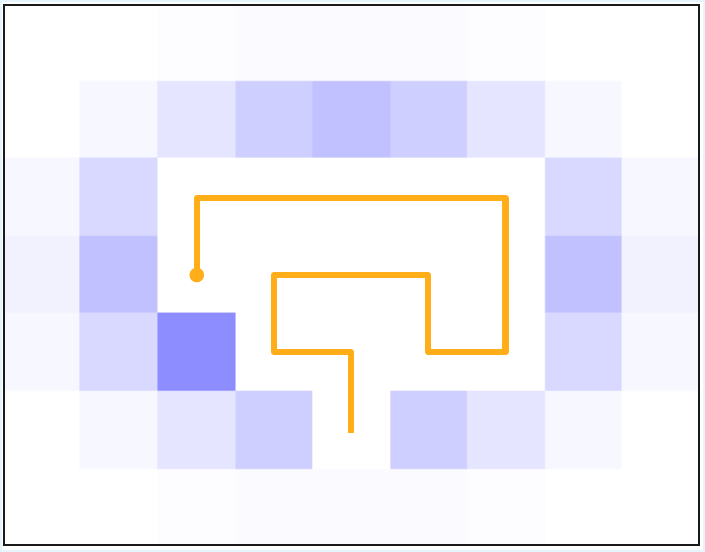}
    \label{fig:3}
    \caption{t = 15}
  \end{subfigure}
  \begin{subfigure}[b]{0.225\linewidth}
    \includegraphics[width=\linewidth]{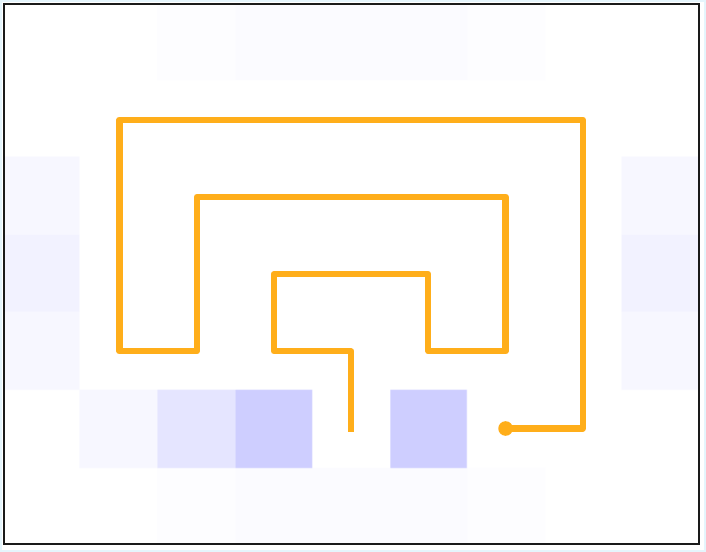}
    \label{fig:4}
    \caption{t = 30}
  \end{subfigure}
  \begin{subfigure}[b]{0.059\linewidth}
    \includegraphics[width=\linewidth]{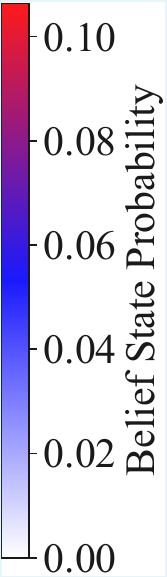}
    \label{fig:4}

  \end{subfigure}

  \caption{\footnotesize The optimal searcher path (orange) calculated by \asar on a toy target distribution for a path length budget of 30, a perfect sensor and fixed starting position. The searcher maintains a target belief state that is updated and plotted with the searcher path. \asar plans a path to optimize the MTTD of a given object probability distribution. The belief state probability is reduced according to the sensor model when \asar searches a vertex.}
  \label{fig:pagetwo}
\end{figure*}

Optimal searcher planning has been studied since the 1940s \cite{Koopman_1946} and is provably NP-hard when maximizing the probability of detection and NP-complete when minimizing MTTD \cite{Trummel_Weisinger_1986}. Stone \cite{Lawrence_1975} develops methods for the optimal distribution of resources between search areas to solve the MTTD problem. Eagle \cite{Eagle_1984} extends these techniques to solve the searcher problem where the path is constrained to adjacent cells as a partially observable Markov decision process (POMDP).
This method only works for small problems (e.g., three-by-three grids) and finds paths that maximize the probability of detection instead of minimizing MTTD.  
Subsequent research in probabilistic search planning has largely continued to focus on the development of methods to maximize detection probability \cite{Stewart_1979, Eagle_Yee_1990, Dell_Eagle_Alves, Bourgault_Furukawa_Durrant-Whyte_2006, Lau_2007, Berger_Lo_2015, Kasthurirangan_Nguyen_Perk_Chakraborty_Mitchell_2025}. A detailed review of probability of detection methods is available in \cite{Raap_Preuß_Meyer-Nieberg_2019}. 

Lau et al. \cite{Lau_Huang_Dissanayake_2005} use dynamic programming to compute an optimal searcher policy that minimizes MTTD for multiple targets in an indoor environment. This method scales poorly, assumes a perfect sensor and stationary targets and does not consider path budget constraints.

Sarmiento et al. \cite{Sarmiento_Murrieta-Cid_Hutchinson_2009} consider the MTTD problem for a robot searching for a target through a building with a uniform probability distribution over the target location. This finds approximate solutions by dividing the search problem in two levels. The high level planner decides the order of room visitation using depth-first search and the low level planner finds the robot-trajectory using the Newton-Raphson method. This method is not globally optimal and does not extend to the general discrete-graph MTTD problem. 

Hollinger et. al. \cite{Hollinger_Singh_Djugash_Kehagias_2009} introduce the Multi-robot Efficient Search Path Planning problem, which approximates the MTTD problem with a time-discounted reward (TDR) objective function. It enumerates all possible paths up to a fixed planning depth then applies sequential allocation that scales linearly with the number of searchers. This finds near-optimal solutions to the finite-horizon TDR problem but provides no guarantees to the MTTD problem.  

The cooperative graph-based model predictive search algorithm  \cite{Riehl_Collins_Hespanha_2011} uses receding-horizon planning and dynamic probability-weighted graphs to plan the searcher paths of multiple UAVs. It achieves faster detection than greedy or parallel-track baselines with proven finite-time discovery guarantees but uses a probability-of-detection and travel-cost reward function that does not guarantee global MTTD optimality. \squeezeWords 

Lanillos \cite{Lanillos_thesis_2013} uses constraint programming (CP) to find solutions to the TDR approximation of the MTTD problem. This finds the optimal TDR solution but provides no guarantees to the MTTD problem. It is also only applicable to small problems and assumes a stationary target and a perfect sensor. 

Lanillos \cite{Lanillos_thesis_2013} also introduces a truncated version of the MTTD function which can be optimized over a finite planning horizon. They use cross entropy optimization (CEO) to find searcher paths that minimize this objective. CEO uses nondeterministic sampling to reduce the complexity and can solve larger problems than the CP method without unrealistic sensor and target assumptions but provides no guarantees on the quality of solutions found in finite time.  

Perez-Carabaza et al. \cite{Pérez-Carabaza_2018} use ant colony optimization (ACO) to find solutions to the MTTD problem, which also uses nondeterministic sampling. They calculate a heuristic to guide the search towards higher areas of target probability. ACO outperforms other approximate methods, including CEO, Bayesian optimization \cite{Lanillos_Yañez-Zuluaga_Ruz_Besada-Portas_2013} and genetic algorithms \cite{Lin_Goodrich_2009} but provides no guarantees on solution quality in finite time. 

A related problem in robotics is informative path planning (IPP), which seeks paths that maximize information collected from the environment given a budget constraint. Typical objectives include minimizing entropy or maximizing mutual information, both of which are order-invariant, unlike minimizing MTTD. Some techniques from IPP can be applied to MTTD searcher path planning despite these differences. 

 Existing solution approaches for IPP include sampling-based planning \cite{Hollinger2014Sampling, Moon_2022}, nondeterministic sampling \cite{Hitz2017Adaptive}, and graph-search methods \cite{Chekuri2005Recursive, Binney2012Branch, Schlotfeldt_Atanasov_Pappas_2019}. Schlotfeldt et al. \cite{Schlotfeldt_Atanasov_Pappas_2019} calculate an admissible and consistent heuristic to inform an A*-style search. The heuristic lower bounds the remaining uncertainty cost by substituting the true sensor information matrix with an upper bound over the robot’s reachable set. This heuristic is only applied to order invariant objectives and time-invariant environments. 

A* \cite{Hart_Nilsson_Raphael_1968} has been applied to searcher path planning. Guo et. al. \cite{Guo_Yi_Hao_2023} use A* to guide a UAV from its launch base to the target search area while avoiding obstacles. A* was not used to find the searcher path inside the search area which was instead calculated using a coverage path planning (CPP) algorithm. Neither A* nor the CPP algorithm optimized for probability of detection or MTTD. 

Alternatively, \asar does not require unrealistic assumptions about the target or sensor and optimizes MTTD directly. Unlike approximate methods, \asar bounds the search with a heuristic calculated from the searcher's reachable set, as in \cite{Schlotfeldt_Atanasov_Pappas_2019}. Unlike \cite{Schlotfeldt_Atanasov_Pappas_2019}, \asar extends the heuristic for an order-{} and time-variant problem. \asar orders its A*-style search for a searcher path with this heuristic to find solutions that are guaranteed to be within a user-specified factor, $\varepsilon$, of the optimal solution, unlike \cite{Guo_Yi_Hao_2023}.  
 
\section{A* For Search and Rescue (\asar)}
\asar is a graph search algorithm based on A* \cite{Hart_Nilsson_Raphael_1968} that finds $\varepsilon$-optimal searcher paths to minimize the finite-path MTTD objective function given an initial position, searcher path budget, prior probability distribution over target location, target movement and sensor models.

A* is a best-first search algorithm ordered by potential solution quality that finds minimum-cost paths in a graph. It extends Dijkstra's algorithm \cite{Dijkstra_1959} to evaluate vertices in order of potential solution quality as given by a heuristic estimate of the cost of a solution path passing through the vertex. If this heuristic never over estimates the true solution cost, then it is said to be admissible and A* is guaranteed to find the optimal solution as efficiently as any other equivalent algorithm, in terms of the number of vertex expansions \cite{Hart_Nilsson_Raphael_1968}. 

Weighted A* \cite{Pohl_1970} introduces an inflation parameter, $\varepsilon > 1$, to the heuristic. This biases the search towards the goal and often expands fewer vertices than A* while sacrificing optimality guarantees. Weighted A* is guaranteed to find a suboptimal solution no worse than the inflation factor times the optimum. \squeezeWords

\asar applies A* to minimize the MTTD of a path through a belief state that evolves both with time and the route taken (i.e., time and order \emph{variant}). It does this while considering a maximum path budget by representing the problem as a graph where each vertex couples the searcher position and the belief state of the target location. 

\asar maintains a priority queue of vertices ordered by summing the MTTD objective function to reach a vertex (i.e., objective-to-come) and a heuristic estimate of the remaining objective available (i.e., objective-to-go). The vertex with minimum value is removed from the queue at each iteration and its outgoing edges are expanded (i.e., its child vertices are added to the queue). \asar does not have a specific goal and the goal is instead defined as any vertex that has an incoming path with length equal to the budget. 

\asar also weights the heuristic by an inflation factor, $\varepsilon \geq 1$, to prioritize expanding vertices near a goal. This often expands fewer vertices while guaranteeing that the returned path is within the inflation factor of the optimum, as in Weighted A*. This parameter allows the user to balance optimality and computational efficiency. \squeezeLine

\subsection{Notation}
Probabilistic searcher planning often represents the location of the target with a discrete probability map where each element stores the likelihood that it contains the target. The map is modelled as a finite, graph, $G = (V, E)$, where $V$ is the set of vertices and $E \subseteq V \times V$ is the set of edges, where a bidirectional edge exists between all spatially adjacent vertices (e.g., 4- or 8-connected).

The initial probability distribution over the target location is denoted as $p$. The search for the target occurs in discrete time and over a finite time horizon, $t \in (1, ..., T)$, where $T$ denotes the searcher path budget. 

The searcher can only search for the target in one vertex at each time step. The searcher path, $\sigma$, is a sequence of edges, $(v_i, v_j) \in E$, such that $\sigma = \left((v_0, v_1), (v_1, v_2), ..., (v_{n-1},v_n)\right)$. A solution is a path such that the number of edges is equal to the searcher path budget, i.e., $n=T$. 

\asar maintains a priority queue, $Q$, of states. Let $A$ be any set and let $B$ and $C$ be subsets of $A$, i.e.,
$B,C \subseteq A$. The notation $B \overset{+}{\gets} C$ is used for $B \gets B \cup C$ and $B\overset{-}{\gets}C$ for $B \gets B \setminus C$. 

\subsection{Problem Preliminaries}
 The belief state that represents the probability that the target is present and has not been detected at a vertex, $v$, at a time, $t$, given the searcher path, $\sigma$, is denoted as $\bel(v, t \mid\sigma)$. Note that the belief state is only affected by subset of the path that occurred before the query time. The belief state is initialized as the initial probability distribution over the target location, $p$, at the beginning of the search, i.e., $\forall v \in V, b(v, 1) = p(v)$. The belief state is not a probability distribution as the total mass in the belief at any time is the probability that the target has yet to be detected \cite{Pérez-Carabaza_2018}.

 Search targets often move during the search (e.g., ocean drift). The motion of the target from one vertex, $v_i$, to another, $v_j$, at a time, $t$, is modelled  by $P\left( v_j \mid v_{i}, t\right)$. The belief state evolves through time via a motion model, 
\begin{equation}
    \ibel(v_j, t \mid\sigma) = \sum_{v_i \in V}P(v_j\mid v_i, t)\bel(v_i, t-1 \mid\sigma),
\end{equation}
where $\ibel$ is the \textit{a priori} belief state that accounts for the target motion to time $t$ but only the search up to time $t-1$. 

The probability that the searcher detects the target at a vertex, $v$, if it is present (i.e., glimpse probability) is given by $\Glimpse{v}$. Glimpse probability is assumed to be independent and identically distributed (i.i.d.) between timesteps and no false-positive detections are considered. When a vertex, $v$, is searched at time, $t$, its \textit{a priori} belief is decreased by the glimpse probability, $\bel(v, t \mid\sigma) = \left(1 - \Glimpse{v}\right)\ibel(v, t \mid \sigma)$. 

We define a state, $x \in X$, as the tuple consisting of a vertex, $v$, the partial path taken to the vertex, $\sigma_x$, and the stored belief state, $b_x[\cdot]$, i.e., $x \coloneqq\{v, \sigma_x, b_x[\cdot]\}$. The cost of a state, denoted as $t(x)$, is the travel time required to reach that state along the path. The stored belief state, $b_x[\cdot]$, is represented as an array with each entry corresponding to the belief value, $\forall u \in V, \bel_x[u] = \bel\!\left(u, t(x) \mid \sigma_x \right)$. 

\subsection{Problem Formulation}

Two common objective functions in probabilistic searcher planning are maximizing the probability of detection, $P_d$, and minimizing the MTTD. The probability of detection of a path of length $T$ is defined as
\begin{equation}
    P_d(\sigma)=\sum_{v\in V}\left(1-\bel(v, T \mid \sigma)\right),
\end{equation}
where $\bel(v, T \mid \sigma)$ is the probability that the target is at vertex $v$ and has not been detected by the searcher path. 

The expected detection time of a path is defined as,
\begin{equation}
\label{ET1}
\begin{aligned}
E[D(\sigma)] &= \sum_{t=1}^{\infty}tP\left(D(\sigma\right)=t) \\ &=\sum_{t=1}^{\infty}t\,\Glimpse{\sigma(t)}\ibel(\sigma(t), t \mid\sigma),
\end{aligned}
\end{equation}
where $D(\sigma)$ is the detection time of the searcher taking the path, $\sigma$, and $\sigma(t)$ is the vertex occupied by the path at time $t$. This objective function requires a sum over all time and when approximated to a finite time results in a suboptimal expected detection time \cite{Lanillos_thesis_2013}. When \eqref{ET1} is rewritten as 
\begin{equation}
\label{ET2}
    E[D(\sigma)] = \sum_{t=1}^{\infty}P(D(\sigma) >t),
\end{equation}
it can be truncated to
\begin{equation}
  \label{ET3}
    J(\sigma) = \sum_{t=1}^{T}P(D(\sigma) >t)= \sum_{t=1}^{T}\sum_{v\in V}\bel(v, t \mid\sigma),
\end{equation}
where $J(\sigma)$ is the finite-path-budget MTTD objective function of a path, $\sigma$, and the sum, $\sum_{v\in V}\bel(v, t \mid\sigma)$, is the probability that the target has not been detected at time $t$ by the searcher following the path \cite{Lanillos_thesis_2013}. A path that minimizes this truncated version is optimal for the path budget, $T$, even though it returns a lower estimate of the expected target detection time \cite{Lanillos_thesis_2013}. Formally the MTTD problem is then to find the optimal path, $\sigma^*$, such that,

\begin{equation}
\label{objective}
    \sigma^* = \argmin_{\sigma} J(\sigma)=\argmin_{\sigma}\sum_{t=1}^{T}\sum_{v\in V}\bel(v, t \mid\sigma).
\end{equation}

The truncated MTTD objective function is non-Markovian since the portion of the objective from a vertex to the end of a path is dependent on the partial path up to that vertex.     

\subsection{Path Budgeted Search (Algorithm \ref{alg:asar})}
\asar maintains a priority queue of states, $Q$, ordered by the $\varepsilon$-weighted evaluation function, $f(x) = \rtc{x} + \varepsilon\rtg{x}$, where $\varepsilon$ is the suboptimality factor, $\rtc{x}$ is the portion of the objective function accrued to reach the state (i.e., the objective-to-come) and $\rtg{x}$ is the admissible heuristic estimate of the objective function to be accrued from the state (i.e., the objective-to-go). The objective-to-come is calculated by summing the objective over the path to the vertex,
\begin{equation}
    \rtc{x} = \sum_{t=1}^{t(x)}\sum_{v\in V}b(v, t \mid \sigma_x). 
\end{equation}
The admissible heuristic estimate of the objective-to-go is always less than the unknown optimal objective-to-go,
\begin{equation}
    \rtg{x} \leq \sum_{t=t(x)+1}^{T}\sum_{v\in V}b(v, t \mid \sigma_x  \oplus \sigma^*), 
\end{equation}
where $\sigma^*$ is the best possible path achievable from the current state with the remaining budget and $\oplus$ denotes the concatenation of two paths. The heuristic estimate of the objective-to-go is calculated using Algorithm \ref{alg:heuristic}.   

\begin{algorithm}[tbp]
\small
\label{asar_alg}
%\SetInd{0.15em}{1em}
\caption{$\rm{A}^{\rm{SAR}}$(\(x_{\rm{start}},  T, \varepsilon\))}\label{alg:asar}

$Q \gets \{x_{\rm{start}}\};\;g_{\rm{best}} \gets \infty; \;x_{\rm{top}} \gets x_{\rm{start}};$ \\
\While{$t(x_{\rm{top}}) < T$}{
    $Q \overset{-}{\gets} \{x_{\rm{top}}\};$ \\
    \For{$(v_{x_{\rm{top}}}, v_{\rm{new}}) \in E$}{
    
    $t_{{\rm{new}}} \gets t(x_{\rm{top}}) + t(v_{\rm{top}}, v_{\rm{new}});$ \\
        \If{$t_{{\rm{new}}} \leq T$}{
        $\sigma_{\rm{new}} \gets \sigma_{x_{\rm{top} }} \oplus (v_{\rm{top}}, v_{\rm{new}});$ \\
            \For{$v\in V$}{
                $\ibel_{\rm{new}}[v] \gets \sum\limits_{u \in V} P\!\left(v \mid u, t_{\rm{new}}\right) b_{x_{\rm{top}}}[u];$
            }
            $b_{\rm{new}}[v_{\rm{new}}]\gets (1-q(v_{\rm{new}}))\ibel_{\rm{new}}[v_{\rm{new}}]  ;\!$ \\
            \For{$v\in \left(V\setminus  v_{\rm{new}}\right)$}{
            $b_{\rm{new}}[v] \gets \ibel_{\rm{new}}[v];$
            }
            $x_{\rm{new}} \gets \{v_{\rm{new}}, \sigma_{\rm{new}}, b_{\rm{new}}[\cdot] \};$ \\
            \If{$\rtc{x_{\rm{new}}} + \varepsilon\rtg{x_{\rm{new}}} < g_{\rm{best}}$}{
            $Q \overset{+}{\gets} \{x_{\rm{new}}\};$ \\
            
            \If{$\rtc{x_{\rm{new}}} < g_{\rm{best}}\; \operatorname{and}\; t_{\rm{new}} \equiv T$}{
                $x_{\rm{best}} \gets x_{\rm{new}};$ \\
                $g_{\rm{best}} \gets \rtc{x_{\rm{new}}};$ \\
                }
            }
        }     
    }
    $x_{\rm{top}} \gets \operatorname*{argmin}\limits_{x \in Q}  \{\rtc{x} + \varepsilon\rtg{x}\};$
}
\Return $x_{\rm{best}}$
\end{algorithm}
 
The best solution objective function value, $g_{\rm{best}}$, is initially set to infinity (Line 1). The state with minimum $\varepsilon$-weighted evaluation function value (i.e., $f$-value), $x_{\rm{top}} \coloneqq \arg\min_{x\in Q} {f(x)}$, is removed from the queue at each iteration (Lines 3, 19). The successor states corresponding to the current state are then generated and evaluated (Lines 4--18). 

The path cost of each successor (i.e., new) state, $t_{\rm{new}}$, is calculated by adding the path cost of the current state, $t({x_{\rm{top}}})$, and the path cost of the edge between the current state and the successor state, $t(v_{\rm{top}}, v_{\rm{new}})$ (Line 5). If the new state's path cost is less than or equal to the search budget, $t_{\rm{new}} \leq T$, the stored \textit{a priori} belief state, $\ibel_{\rm{new}}[\cdot]$, is constructed using the motion model, $P(v_i\mid v_j, t)$ (Lines 8--9). The stored belief state, $\bel_{\rm{new}}[\cdot]$, is then constructed from the \textit{a priori} belief state and by accounting for the effect of searching the new vertex, $v_{\rm{new}}$ (Lines 10--12).

If the $f$-value of the successor state is less than the best solution, $f(x_{\rm{new}}) = \rtc{x_{\rm{new}}} + \varepsilon\rtg{x_{\rm{new}}} < g_{\rm{best}}$, then it could possibly improve the solution and the state is added to the queue, $Q$ (Lines 14--15). The addition of this check reduces memory usage by only adding states that could possibly improve the solution to the queue and does not affect the solution since any state that fails the check would never be removed from the queue before the terminal state. The best solution, $g_{\rm{best}}$, is only updated when a goal state is added to the queue with a lower objective-to-come than the current best solution (Lines 16--18). A goal vertex is any vertex that uses the entire path budget, i.e., $t_{\rm{new}} = T$. \squeezeWords

\asar ends when a goal vertex is removed from the queue, as in A*.  The first goal vertex removed from the queue will have an $f$-value equal to the best solution objective function value, i.e., $f(x_{\rm{goal}}) = g_{\rm{best}} + 0$, and every vertex left in the queue must have a higher $f$-value, $\forall x \in Q,  g_{\rm{best}} \leq g(x) + \varepsilon \rtg{x}$. The admissible heuristic never overestimates the objective-to-go, i.e., $\forall x, g(x) + \rtg{x} \leq g(x) + h(x)$, which ensures that the best solution’s objective function value is guaranteed to be within the suboptimality factor, $\varepsilon$, of the optimal solution, as in Weighted A*.

\subsection{Heuristic Calculation (Algorithm 2)}

\asar calculates a heuristic estimate of the objective-to-go in a manner similar to \cite{Schlotfeldt_Atanasov_Pappas_2019} to order the search by solving a relaxed problem where at each future step, $t>t(x)$, the searcher may search any vertex reachable within $t-t(x)$ steps from the current location without being constrained by path continuity. The resulting sequence of vertices is a lower bound on the path-constrained MTTD objective function (Theorem \ref{thm:proof}). It is calculated by greedily selecting the vertex that maximizes the probability of detecting the target, $q(v)b[v]$,  at each time step.

The heuristic value, $\hat{h}$, is initially set to zero (Line 1). The stored heuristic belief state, $b[\cdot]$, is initialized as the stored belief state of the inputted state, $b_x[\cdot]$ (Lines 2--3). The heuristic is calculated by iterating through each time step in the remaining budget i.e., $t \in ( t(x) +1, t(x) +2, ..., T)$ (Line 4). At each time step, the \textit{a priori} belief state, $\ibel[\cdot]$, is constructed using the motion model, $P(v_i\mid v_j, t)$ (Lines 5--6). The vertex with the highest probability of detection that is reachable from the current vertex is then found (Line 7). The stored heuristic belief state, $\bel[\cdot]$, is finally updated to include the effect of searching the selected vertex (Lines 8--10). The heuristic value is updated by calculating the probability that the target has not been detected by time $t$ and appending it to the running value (Line 11). The final heuristic value is returned after iterating though all time steps up to the path budget, $T$ (Line 13).

\begin{theorem}[Admissibility]

\label{thm:proof}
The estimate of the objective-to-go, $\hat{h}(\cdot)$, never overestimates the optimal objective-to-go, $h^*(\cdot)$, and is admissible, i.e.,
\[
\forall x\in X, \;\rtg{x} \leq h^*(x).
\]
\end{theorem}
\begin{proof*}
Define a relaxation in which at each future step, $t>t(x)$, the searcher may search 
any vertex reachable within $t-t(x)$ steps from the current location without considering path continuity. The optimal value of this relaxed problem underestimates the optimal objective to go, 
\begin{equation}
    h^*_{\mathrm{relax}}(x) \;\le\; h^*(x), 
\end{equation}
because all possible solutions to $h(x)$ lie within the set of possible solutions to $h_{\mathrm{relax}}(x)$. 

The probability mass of the target location is preserved through time and only lowered when the searcher searches a vertex, 

\begin{equation}
    \sum\limits_{v\in V} \ibel(v,t + 1) = \sum\limits_{v\in V} \bel(v,t),
\end{equation}
\begin{equation}
    \sum\limits_{v\in V}\bel(v, t) = \sum\limits_{v\in V}\ibel(v, t) - q(u)\ibel(u, t).
\end{equation}

If there exists some larger reduction in probability, $q(v)b(v,\tau) > q(v)b(v, t)$, at some later time, $\tau > t$, then exchanging the searches so that larger reduction in probability occurs first reduces (or leaves
unchanged) all future probability and strictly reduces the objective. The optimal solution to the relaxed problem is therefore greedy; at each step, $t$, choose the vertex maximizing the reduction in probability mass, $q(v)\ibel(v, t)$. Algorithm \ref{alg:heuristic} implements this greedy solution, and the heuristic value is therefore the optimal value of the relaxed problem,  

\begin{equation}
    \rtg{x} =  h^*_{\mathrm{relax}}(x).
\end{equation}

Since the optimal solution to the relaxed problem lower bounds the optimal objective-to-go, $h^*_{\mathrm{relax}}(x) \leq h^*(x)$, the heuristic is admissible, $\rtg{x}\leq h^*(x)$. 
\end{proof*}
\begin{algorithm}[tbp]
\small
\caption{Heuristic$\left(x\right)$}\label{alg:heuristic}

$\hat{h} \gets 0;$\\ 
\For{$v \in V$}{
$b[v]\gets b_x[v];$
}
\For{$t \in \left(t(x)+1, t(x) +2, ..., T\right)$}{
    \For{$v\in V$}{
    $\ibel[v] \gets \sum\limits_{u \in V}P(v \mid u, t)\bel[u]$
                }
    $v_{\rm{best}} \!\gets\! \operatorname*{argmax}\limits_{v \in V}\{q(v)\ibel[v] \mid t(v_x, v) \leq t \};$ \\
    $\bel[v_{\rm{best}}] \gets (1-q(v_{\rm{best}}))\ibel[v_{\rm{best}}];$\\
    \For{$v\in (V\setminus  v_{\rm{best}})$}{
                $\bel[v]\gets \ibel[v]$
                }
    $\hat{h} \gets \hat{h} + \sum\limits_{v \in V}b[v]$
    
}
\Return{$\hat{h}$}
\end{algorithm}

\section{Experiments}
\newcommand{\spaceabovesubcaption}{-\baselineskip}
\newcommand{\spacebelowsubcaption}{\baselineskip}
% In the document body:
\begin{figure*}[t]      % “*” makes it double‐column; “t” places it at the top
  \centering
  %% Top row
  \begin{subfigure}[b]{0.45\linewidth}
    \includegraphics[width=\linewidth]{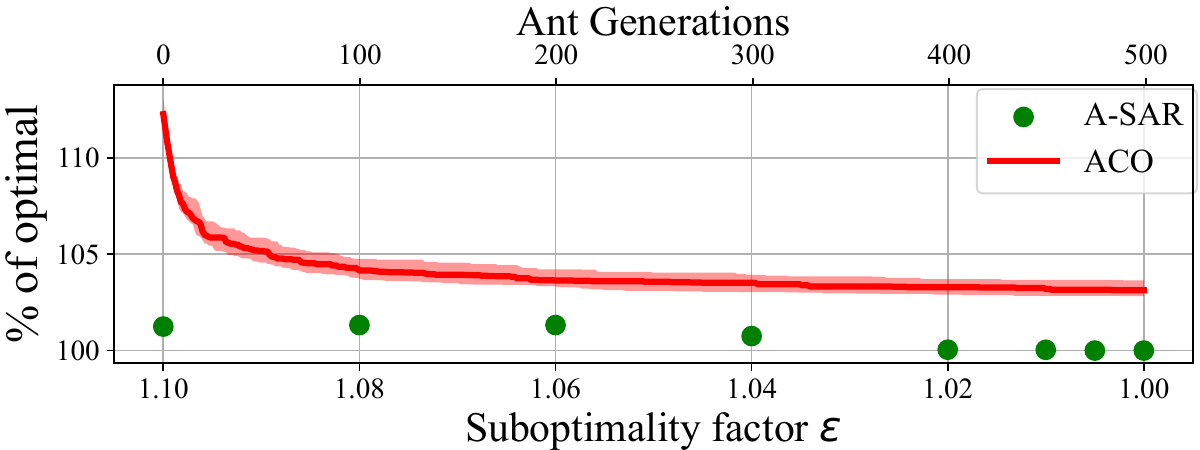}
    \caption{Bay of Fundy, Parallel Track=134\%}
    \label{fig:1}
  \end{subfigure}\quad%
  \begin{subfigure}[b]{0.45\linewidth}
    \includegraphics[width=\linewidth]{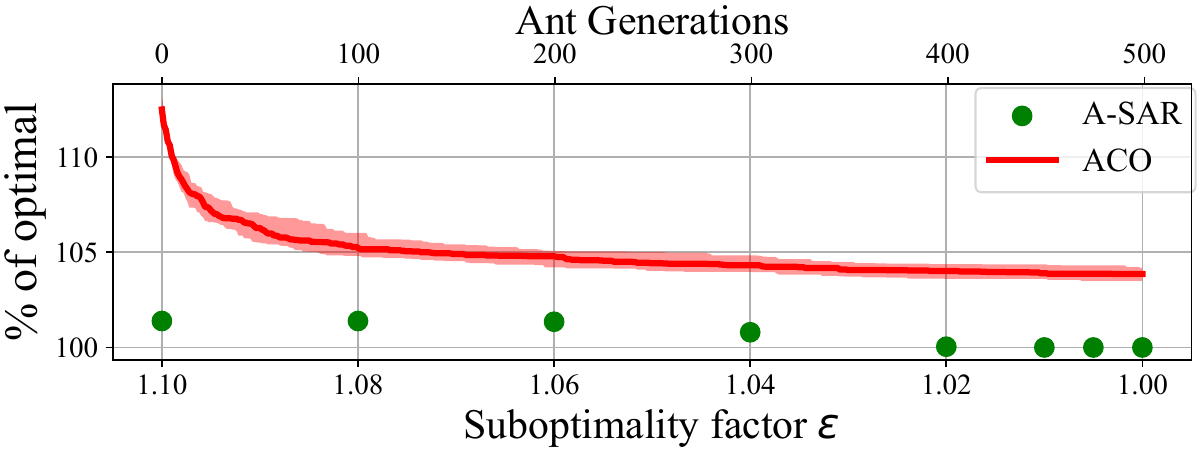}
    \caption{Lake Ontario, Parallel Track=153\%}
    \label{fig:2}
  \end{subfigure}\\%
  %\vspace{-2ex}         % small vertical gap between rows
  %% Bottom row
  \begin{subfigure}[b]{0.45\linewidth}
    \includegraphics[width=\linewidth]{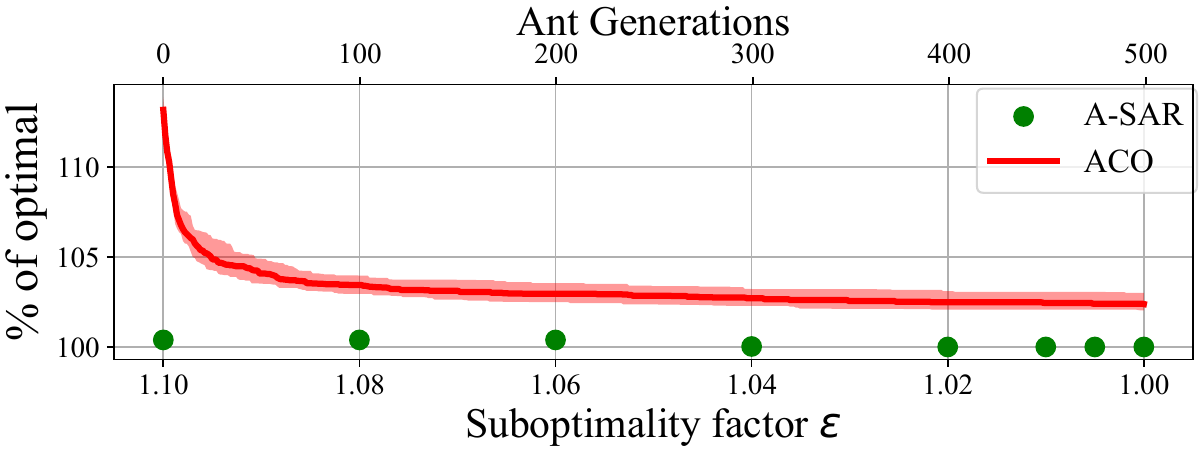}
    \caption{Salish Sea, Parallel Track=152\%}
    \label{fig:3}
  \end{subfigure}\quad%
  \begin{subfigure}[b]{0.45\linewidth}
    \includegraphics[width=\linewidth]{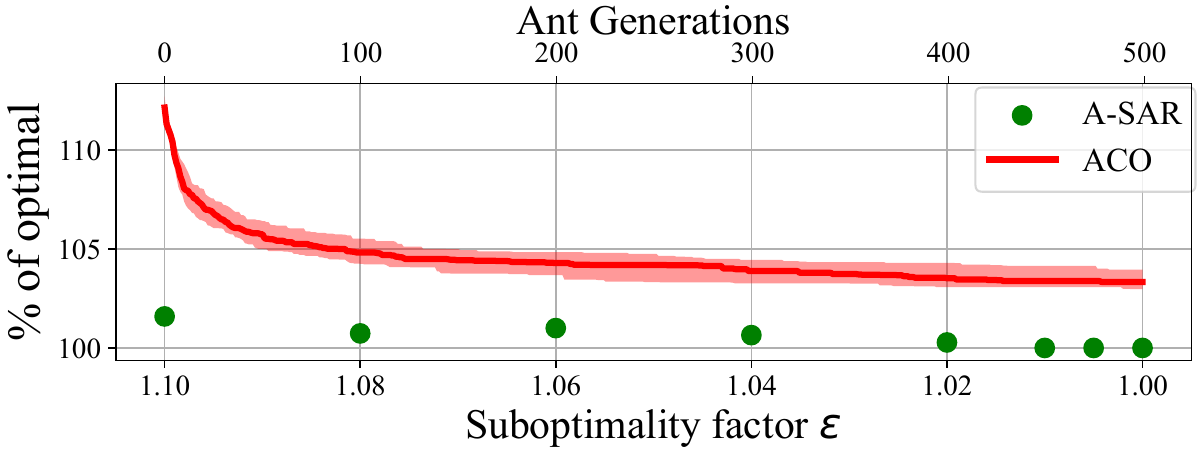}
    \caption{Arctic Ocean, Parallel Track=118\%}
    \label{fig:4}
  \end{subfigure}

  \caption{\footnotesize
Experimental results for Bay of Fundy (\ref{fig:1}), Lake Ontario (\ref{fig:2}), Salish Sea (\ref{fig:3}) and Arctic Ocean (\ref{fig:4}) scenarios comparing an approach from \cite{Pérez-Carabaza_2018} and \asar. The results are expressed as a percentage of the optimal solution. The median values of the ACO planner after 100 trials are plotted along the ant generations axis with nonparametric 99\% confidence intervals. The results of \asar are plotted along the suboptimality factor, $\varepsilon$, axis. The parallel track (PT) baseline results for each scenario are included in the figure labels. \asar outperforms ACO and parallel track in all tested scenarios for all suboptimality factors while providing optimality guarantees.}
  \label{fig:four}
\end{figure*}
The \asar algorithm was developed as part of a project focusing on maritime SAR and experiments were conducted in the maritime domain.

\subsection{Drift Model} 
SAR planning professionals frequently employ Monte Carlo particle-drift models to predict object movement~\cite{Kratzke_Stone_Frost_2010, open_drift}, which are then used to plan searches for missing vessels or objects at sea.
We generated realistic test scenarios with OpenDrift \cite{open_drift}, an open-source framework for modelling oceanic drift.
We used its Leeway model, derived from the US Coast Guard’s SAR model \cite{Allen2005,AllenPlourde1999,Breivik2012}, which uses empirical coefficients to describe how objects react to wind and current.
By combining these coefficients with environmental data, OpenDrift produces trajectory ensembles that define the initial belief state of object location, $p$. In our implementation, particles define the belief state. Each particle carries an associated probability that it has not been detected (PND).
A particle's PND is reduced by the glimpse probability when the searcher path passes over its cell.
The belief of a cell is obtained by summing the PNDs of all particles it contains.

\subsection{Simulated Experiments}

We used OpenDrift to create four distress scenarios which are summarized in Table \ref{tab:scenarios}. These scenarios occur in the Bay of Fundy (Atlantic Ocean), Lake Ontario (North American Great Lakes), Salish Sea (Pacific Ocean) and Hudson's Bay (Arctic Ocean). Four different OpenDrift search search objects were considered: a person in water (PIW) in an unknown state (PIW-1), a conscious PIW positioned vertically and wearing a PFD (PIW-2), a fishing vessel and a deep-ballast life raft with unknown capacity and loading. These parameters were selected to ensure that \asar was assessed across a broad spectrum of realistic maritime SAR conditions.  

The position and time of the distress (i.e., the last known position; LKP) is specified in all scenarios. Table \ref{tab:scenarios} contains the LKP, distress time, commence search point (CSP), commence search time (CST), wind and current models used in the drift model, sweep width and the path budget. The difference between distress time and CST represent the drift time of the object and vary for the four different scenarios. The sweep width is the spacing between search legs in a parallel track search and is determined based on the sensor model, search object and weather conditions \cite{GovCan_CAMSAR_2014}. 

The probability map for each scenario was created by overlaying a grid on the drift plot and counting the number of particles in each cell for each time step. The cell size was chosen as the appropriate sweep width for the search object and visibility, given that the search was conducted from fixed wing aircraft at 300 ft, travelling at 20 m/s \cite{GovCan_CAMSAR_2014}. These searcher parameters were chosen as they best match the UAV used in the field trial in Section \ref{sec:field}. We use a sensor model with the glimpse probability being 78\% for all vertices, i.e., $\forall v \in V, q(v) = 0.78$, in our experiments \cite{GovCan_CAMSAR_2014}, although any sensor model could be used with \asar.

\begin{figure}[tbp]
  \centering

  % Left block: 2x2 grid
  \begin{minipage}[b]{0.8\linewidth}
    \centering
    
    % Top row
    \begin{subfigure}[b]{0.47\linewidth}
      \centering
      \includegraphics[width=\linewidth]{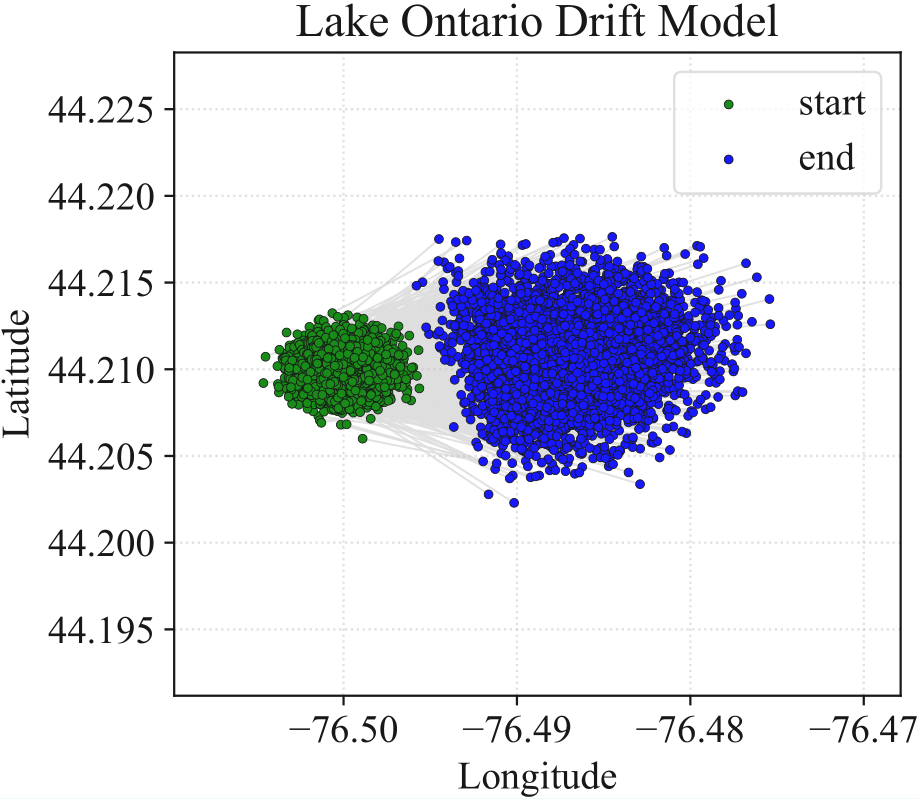}
      \caption{Drift Plot}
      \label{fig:icra_sub1}
    \end{subfigure}
    \hfill
    \begin{subfigure}[b]{0.48\linewidth}
      \centering
      \includegraphics[width=\linewidth]{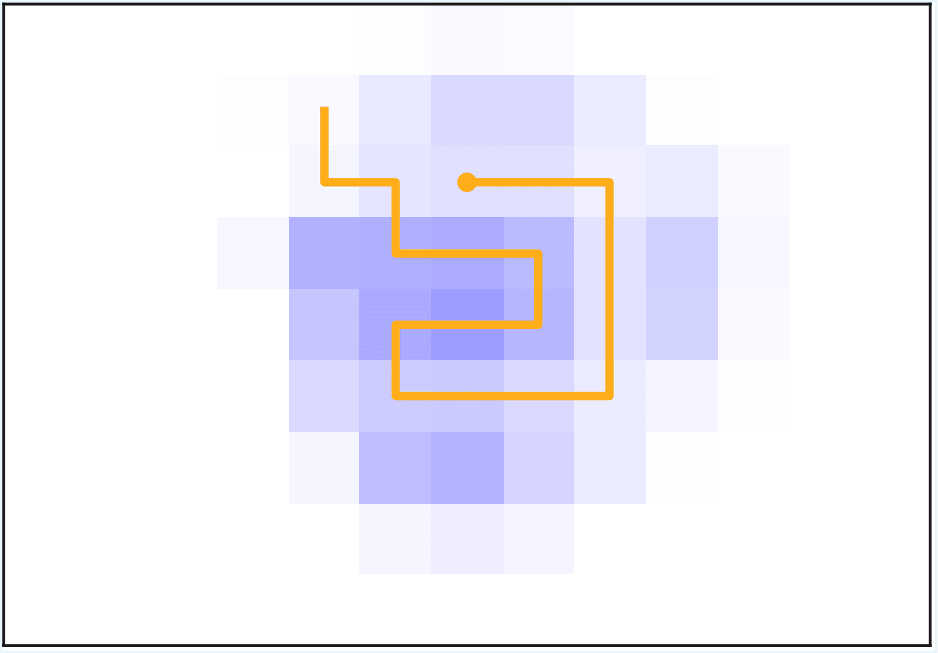}
      \caption{\asar t=19}
      \label{fig:icra_sub2}
    \end{subfigure}
    
    \vspace{1ex}
    
    % Bottom row
    \begin{subfigure}[b]{0.48\linewidth}
      \centering
      \includegraphics[width=\linewidth]{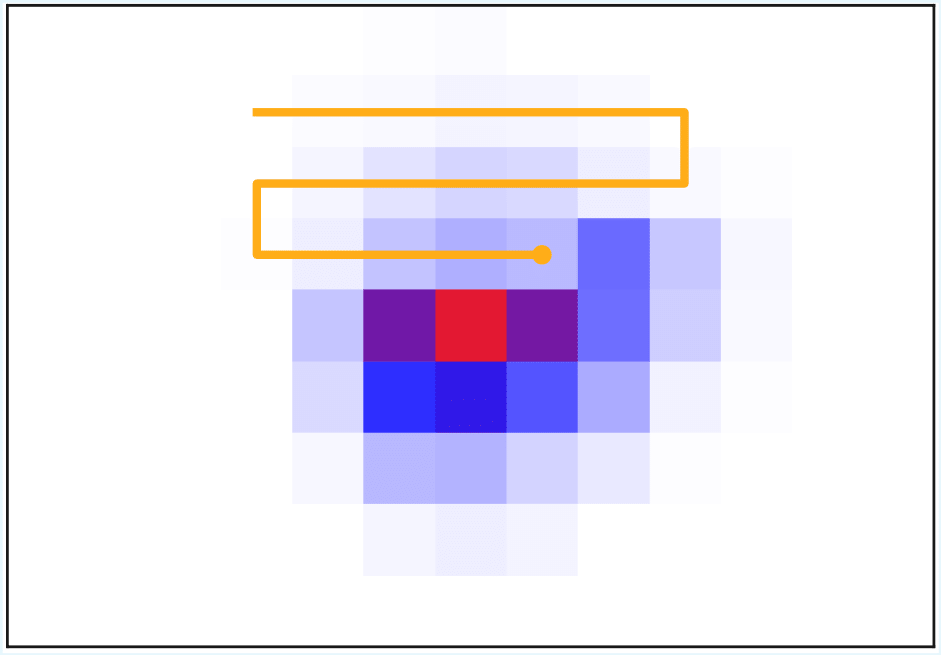}
      \caption{Parallel Track t=19}
      \label{fig:icra_sub3}
    \end{subfigure}
    \hfill
    \begin{subfigure}[b]{0.48\linewidth}
      \centering
      \includegraphics[width=\linewidth]{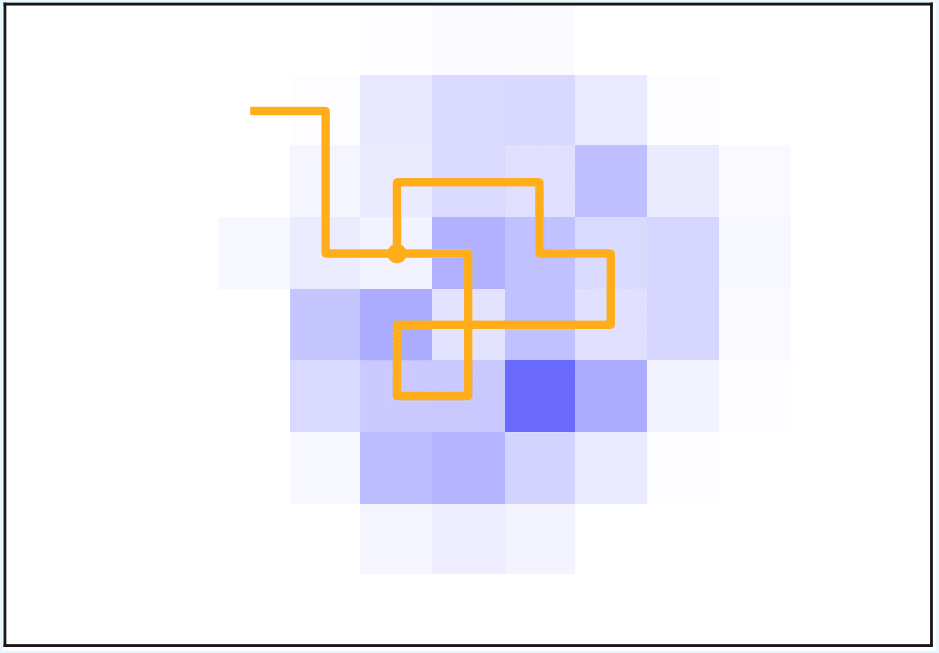}
      \caption{ACO t=19}
      \label{fig:icra_sub4}
    \end{subfigure}
  \end{minipage}%
  \hfill
  % Right block: single tall figure
  \begin{minipage}[b]{0.146\linewidth}
    \centering
    \includegraphics[width=\linewidth]{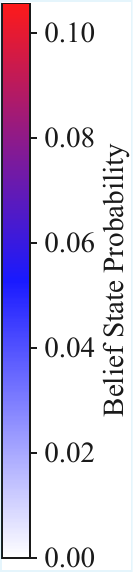}
    \label{fig:icra_sub5}
  \end{minipage}

  \caption{\footnotesize The OpenDrift model and the paths found by \asar, an ACO approach \cite{Pérez-Carabaza_2018} and parallel track on the Lake Ontario scenario. Fig \ref{fig:icra_sub1} shows the OpenDrift Monte Carlo particle drift used to model object location. Figs. \ref{fig:icra_sub2}, \ref{fig:icra_sub3}, and \ref{fig:icra_sub4} show the \asar, parallel track, and ACO searcher paths after 19 steps, respectively. \asar outperforms parallel track and ACO by prioritizing areas with high probability first and conducting the search efficiently, avoiding unnecessary revisits to recently explored vertices when moving toward other high-probability regions.}
  \label{fig:icra_four_subfigs}
\end{figure}

We compared the performance of \asar to the max-min ACO variant in \cite{Pérez-Carabaza_2018}. We implemented the node encoding with the best parameters from \cite{Pérez-Carabaza_2018}. 

\label{sec:field_experiment}
\begin{table*}[tbp]
  \centering
  \setlength{\tabcolsep}{5pt}
  \renewcommand{\arraystretch}{1.3}
  \caption{Testing Scenarios}
  \label{tab:scenarios}
  \begin{tabular}{c c c c c c c c c c}
    \toprule
    \textbf{Scenario} & \textbf{Search Object} 
      & \textbf{LKP} & \textbf{Distress Time} 
      & \textbf{CSP} & \textbf{CST} 
      & \textbf{Wind Model} & \textbf{Current Model} 
      & \textbf{Sweep Width} & \textbf{$T$} \\
    \midrule
    Bay of Fundy   & PIW-1              & \makecell{N\,45.2 \\ W\,65.3} 
                   & \makecell{2025-02-14 \\ 00:00 UTC} 
                   & \makecell{N\,45.283 \\ W\,65.072} 
                   & \makecell{2025-02-14 \\ 09:00 UTC} 
                   & HRRR \cite{hrrr} & GoMOFS \cite{Yang_Richardson_Chen_Kelley_Myers_Aikman_Peng_Zhang_2016} & 0.4\,Nm & 55 \\
    Lake Ontario   & PIW-2              & \makecell{N\,44.21 \\ W\,76.5} &\makecell{2025-08-26 \\ 14:45 UTC} 
                   & \makecell{N\,44.215 \\ W\,76.496} & \makecell{2025-08-26 \\ 15:45 UTC} 
                   & HRRR \cite{hrrr} & LOOFS \cite{Chu_Kelley_Mott_Zhang_Lang_2011} & 0.1\,Nm & 49 \\
    Salish Sea     & Fishing Vessel   & \makecell{N\,48.3 \\ W\,123.07} & \makecell{2025-08-24 \\ 00:00 UTC} 
                   & \makecell{N\,48.195 \\ W\,123.187} & \makecell{2025-08-24 \\ 11:00 UTC}  
                   & HRRR \cite{hrrr} & \makecell{CIOPS \\ Salish Sea \cite{MacDermid2024}} & 1.6\,Nm & 41 \\
    Arctic Ocean   & \makecell{Deep Ballast \\ Life Raft}   & \makecell{N\,62.85 \\ W\,90.6} & \makecell{2025-08-29 \\ 00:00 UTC}  
                   & \makecell{N\,62.786 \\ W\,90.447} &  \makecell{2025-08-30 \\ 00:00 UTC}   
                   & GFS \cite{NCEP2004GFS} & ESPC-D-V02 \cite{epsc_2}      & 0.5\,Nm & 60 \\
    \bottomrule
  \end{tabular}
\end{table*}

We restrict motion to the four cardinal neighbours, excluding diagonals (i.e., 4 connected). This reflects the fixed-wing UAV application where the cell width is the sweep width of a downward-facing camera and diagonal motion would leave parts of the cell unobserved. 

The results of the experiments are plotted in Figure \ref{fig:four}. A baseline was set using a parallel track search created by a certified maritime search planner for each scenario. We then ran \asar and the ACO method from the same starting position and with the same budget as the length of the parallel track search. A plot of the Lake Ontario scenario's modelled drift and resulting searcher paths is shown in Figure \ref{fig:icra_four_subfigs}. 

The performance of the ACO method was measured according to the number of ant generations, a number that increases with time, instead of computational time because our implementation of the ACO was not performance optimized. We plot the median solution quality with nonparametric 99\% confidence interval of 100 trials in each experiment. \asar is deterministic and finds one result for any given problem and suboptimality factor, $\varepsilon$, with solution quality increasing as the suboptimality factor decreases. The solution quality is plotted as a function of the suboptimality factor alongside the ACO results. \asar's wall clock time is presented in Table \ref{tab:time} for each scenario and suboptimality factor. \asar was implemented in C++ and run on a 3.2 GHz M1 Pro processor. \squeezeWords

\subsection{Lake Ontario Field Experiment}
\label{sec:field}
\begin{table}[tbp]
\centering
\setlength{\tabcolsep}{4pt}
\renewcommand{\arraystretch}{1.3}
\caption{\asar Execution Time (seconds)}
\label{tab:time}
\begin{tabular}{lcccccccc} % 9 columns total
\toprule
\multirow[b]{2}{*}{\textbf{Scenario}} & \multicolumn{8}{c}{\textbf{Suboptimality factor $\varepsilon$}} \\
%\cmidrule(lr){2-9}
 & 1.1 & 1.08 & 1.06 & 1.04 & 1.02 & 1.01 & 1.005 & 1.0 \\
\midrule
Bay of Fundy & 0.10 & 0.11 & 0.12 & 0.17 & 0.29 & 1.97 & 5.45 & 19.73 \\
Lake Ontario  & 0.09 & 0.09 & 0.10 & 0.13 & 0.19 & 0.30 & 0.52 &  0.86 \\
Salish Sea & 0.05 & 0.05 & 0.07 & 0.22 & 0.75 & 1.32 & 1.70 & 2.24 \\
Arctic Ocean & 0.10 & 0.10 & 0.10 & 0.50 & 0.52 & 3.09 & 10.78 & 35.26 \\
\bottomrule
\end{tabular}
\end{table}
A field experiment was conducted on Lake Ontario to validate the performance of \asar on a real-world search object. The experiment involved dropping a manikin in position N44.210, W76.500 on 2025-09-08 at 19:15 UTC. The manikin was then allowed to drift freely for one hour. A drift model was created for the manikin using OpenDrift and the same wind and current models as the Lake Ontario scenario in Table \ref{tab:scenarios}. This drift model was used to create the probability map and motion model in the same manner as the previous experiments. \asar was then run to create a searcher path which was executed by a fixed wing UAV (Figure \ref{fig:place_holder}). This UAV is equipped with downward facing RGB and infrared cameras with a sweep width of 0.1 Nm. Figure \ref{fig:place_holder} shows the drift model, drift track, searcher path and point of intersection of the manikin and searcher path of the experiment. \squeezeWords   

\section{Discussion}

Figure \ref{fig:four} shows that \asar finds a better solution in terms of MTTD than the ACO planner in \cite{Pérez-Carabaza_2018} and the parallel-track patterns.
The \(\varepsilon\!=\!1.0\) configuration achieved truncated MTTD objective values that were 3.18\% lower on average than the median solutions produced by ACO and 39\% lower on average than the parallel tracks searches.
Even the weighted setting, \(\varepsilon\!=\!1.1\), produced a solution less than 
$101.5\%$ of the optimal truncated MTTD objective on average and outperformed the median ACO solution in all experiments.

\asar finds these better solutions faster than ACO and provides formal solution quality guarantees. Our implementation of ACO was not performance optimized, so exact ACO run times are not included, but the experimental section of \cite{Pérez-Carabaza_2018} runs ACO for 50s before convergence on problems with significantly shorter path budgets. The number of ants per generation is proportional to path budget and the time required for ant to construct a path increases with path budget. This implies that significantly more computational time would be required for convergence on real scenarios. \asar can find the optimal path in less time and found paths within 1.15\% of optimum on average within 0.1 seconds. This shows that \asar can be used in real-world scenarios. 

The Lake Ontario field trial confirms that this performance transfers to a real-world search target.
The fixed-wing UAV located the drifting manikin in 150 seconds, within 20\% of the detection time forecast by the planner despite localization uncertainty.

%%%%%%%%%%%%%%%%%%%%%%%%%%%%%%%%%%%%%%%%%%%%%%%%%%%%%%%%%%%%%%%%%%%%%%%%%%%%%%%%
\section{Conclusion}
This work introduced \asar{}, an A*-style graph-search algorithm that
computes $\varepsilon$-optimal MTTD searcher trajectories under path budget constraints in dynamic environments. \asar bounds the search space with an admissible heuristic that provides a lower bound on the MTTD objective function. It then uses a best-first search ordered by potential solution quality to find a solution that is within a user-defined suboptimality factor, $\varepsilon$, of the optimal path. \squeezeLine

Simulated experiments using OpenDrift models showed that \asar finds better solutions faster than a leading MTTD searcher planning method and significantly outperforms traditional search patterns. A field trial validated \asar's performance on a real-world search object. Future work will include anytime planning/replanning, multiagent coordination and the inclusion of searcher kinematic constraints.


%%%%%%%%%%%%%%%%%%%%%%%%%%%%%%%%%%%%%%%%%%%%%%%%%%%%%%%%%%%%%%%%%%%%%%%%%%%%%%%%

%%%%%%%%%%%%%%%%%%%%%%%%%%%%%%%%%%%%%%%%%%%%%%%%%%%%%%%%%%%%%%%%%%%%%%%%%%%%%%%%

%%%%%%%%%%%%%%%%%%%%%%%%%%%%%%%%%%%%%%%%%%%%%%%%%%%%%%%%%%%%%%%%%%%%%%%%%%%%%%%%

\bibliographystyle{IEEEtran}
\bibliography{references}

\end{document}